\documentclass{article}
\usepackage{ijcai23}
\usepackage{times}
\usepackage{soul}
\usepackage{url}
\usepackage[hidelinks]{hyperref}
\usepackage[utf8]{inputenc}
\usepackage[small]{caption}
\usepackage{graphicx}
\usepackage{amsmath}
\usepackage{amsthm}
\usepackage{booktabs}
\usepackage{algorithm}
\usepackage{algorithmic}
\usepackage[switch]{lineno}
\usepackage{amssymb}

\title{Team Intro\_to\_AI\_team8 at CoachAI Badminton Challenge 2023: Advanced ShuttleNet for Shot Predictions}

\author{
Shih-Hong Chen
\and
Pin-Hsuan Chou\and
Yong-Fu Liu\And
Chien-An Han
\affiliations
National Yang Ming Chiao Tung University, Hsinchu, Taiwan
\emails
\{tonychen.sc09, sherrychou.sc09, cies96035.ee09, guaninld.ee07\}@nycu.edu.tw
}

\begin{document}

\maketitle

\begin{abstract}
    In this paper, our objective is to improve the performance of the existing framework ShuttleNet in predicting badminton shot types and locations by leveraging past strokes. We participated in the CoachAI Badminton Challenge at IJCAI 2023 and achieved significantly better results compared to the baseline. Ultimately, our team achieved the first position in the competition and we made our code available. \footnote{\url{https://github.com/Collector-77/Intro_to_AI_team8-CoachAI-Badminton-Challenge/tree/main}}
\end{abstract}

\section{Introduction} Badminton is a popular sport that requires agility, skill, and physical strategic decision-making. A crucial aspect of the game is predicting the landing point of the shuttlecock and identifying the opponent's shot type. Accurate prediction can provide players with a significant advantage, enabling them to anticipate and respond effectively to the incoming shots.

In recent years, there has been growing interest in employing data-driven approaches to enhance player performance across various sports, including badminton. However, predicting the landing point and shot type of a shuttlecock is a complex task due to factors such as player positioning, shot technique, and environmental conditions. Traditional approaches to predicting these outcomes have relied on manual observation and subjective judgment, which can be susceptible to errors and inconsistencies. Despite these challenges, advancements in technology and machine learning algorithms offer an opportunity to develop more accurate and reliable predictive models for badminton.  For instance, ~\cite{ShuttleNet_AAAI_2022} had proposed a novel framework ShuttleNet to analyze the kinematic features of the shuttlecock's flight and make predictions.
\nocite{ShuttleNet_AAAI_2022}

This study contributes in two primary ways: firstly, we present a comprehensive dataset analysis, identifying key areas for improvement. Secondly, we propose and validate a set of enhancements that enhance the prediction accuracy of the model.

\section{Related works} ~\cite{ShuttleNet_AAAI_2022} introduced a novel approach that takes into account the interactions between the shot type and landing area of the shuttlecock. They project the shot type, landing coordinates, and players into higher-dimensional embeddings, employing a Transformer-based feature extractor to capture the current situation in the rally and the overall playing style of each player. Additionally, they propose the Position-aware Gated Fusion Network to integrate the rally and player contexts. This fusion is achieved through the prediction layer, generating the final output.

\section{Method} 
\subsection{Problem Formulation} Let $R = \{S_r,P_r\}^{\lvert R \rvert}_{R=1}$ denote historical rallies of badminton matches, where the $r$-th rally is composed of a stroke sequence with type-area pairs $S_{r} = (\langle s_{1}, a_{1}\rangle,\cdots,\langle s_{\lvert S_{r}\rvert}, a_{\lvert S_{r}\rvert}\rangle)$ and a player sequence $P_{r} = (p_{1},\cdots,p_{\lvert S_{r} \rvert})$. At the $i$-th stroke, $s_{i}$ represents the shot type, $a_{i} = \langle x_{i}, y_{i} \rangle \in \mathbb{R}^2 $ are the coordinates of the shuttle destinations, and $p_{i}$ is the player who hits the shuttle. We denote Player A as the served player and Player B as the other for each rally in this paper. For instance, given a single rally between Player A and Player B, $P_{r}$ may become $(A,B,...,A,B)$. We formulate the problem of stroke
forecasting as follows. For each rally, given the observed $\tau$ strokes $(\langle s_{i}, a_{i}\rangle)^\tau_{i=1}$ with players $(p_{i})^\tau_{i=1}$, the goal is to predict the future strokes including shot types and area coordinates for the next $n$ steps, i.e., $(\langle s_{i}, a_{i}\rangle)^{\tau+n}_{i=\tau+1}$. In this task, the value of $\tau$ is fixed at 4.

\begin{figure*}[t]
  \centering
  \includegraphics[width=\textwidth]{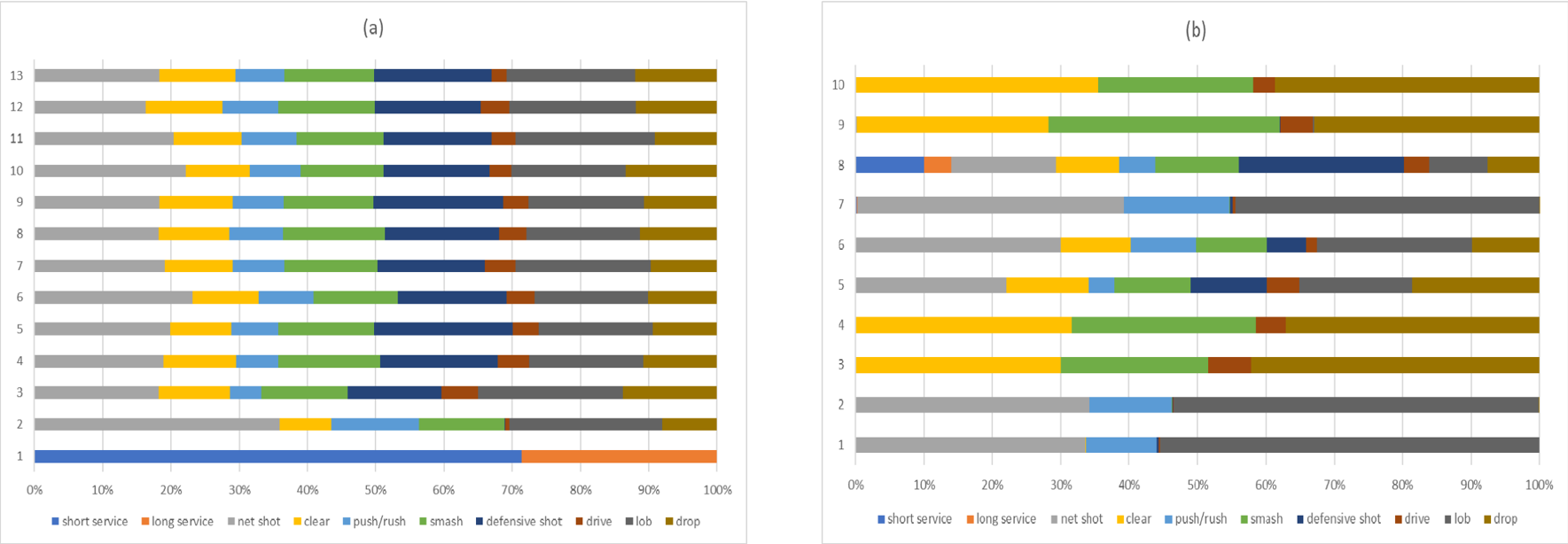}
\end{figure*}

\begin{figure*}[t]
  \centering
  \includegraphics[width=\textwidth]{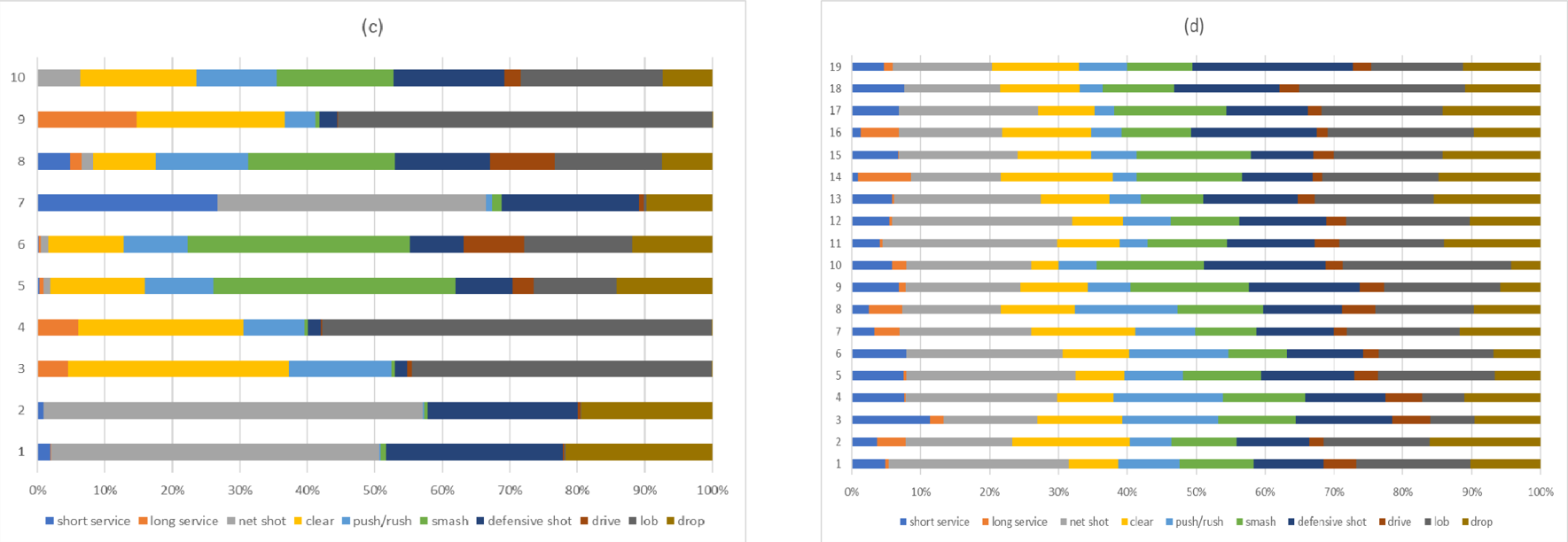}
  \caption*{\begin{minipage}{\textwidth}Figure 1: Distribution of shot types based on (a) ball rounds (partial) (b) player locations (c) landing locations (d) different players (partial) \end{minipage}}
\end{figure*}

\begin{figure*}[t]
  \centering
  \includegraphics[width=\textwidth]{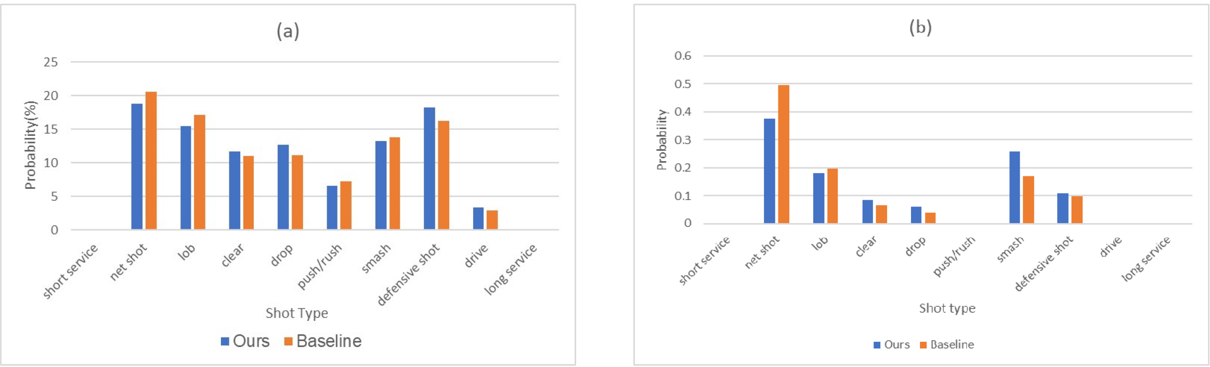}
  \caption*{\begin{minipage}{\textwidth}\centering Figure 2: (a) Probability distribution of predicted shot types (b) Distribution of final predicted shot types \end{minipage}}
\end{figure*}

\begin{figure*}[t]
  \centering
  \includegraphics[width=\textwidth]{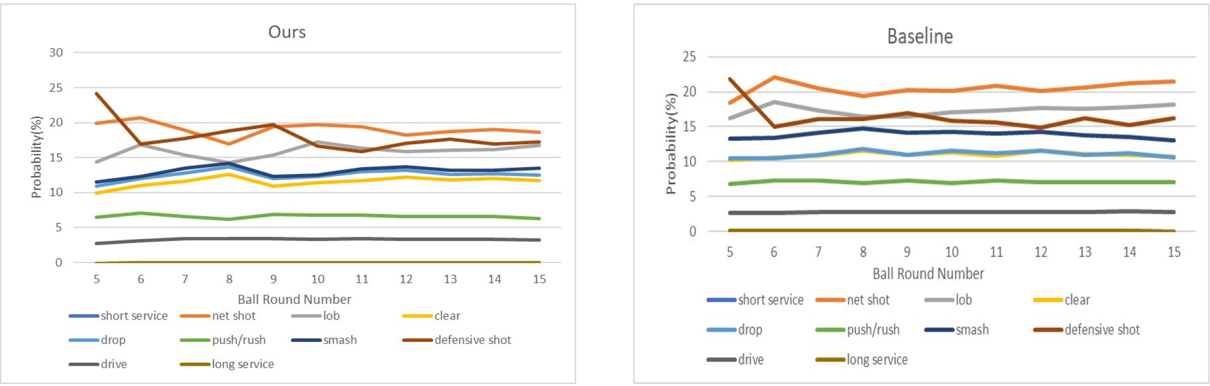}
  \caption*{\begin{minipage}{\textwidth}\centering Figure 3: Trend of prediction probabilities for original ShuttleNet and our modified version across different ball rounds \end{minipage}}
\end{figure*}

\renewcommand{\arraystretch}{1.5}
\begin{table*}[t]
    \begin{minipage}[t]{\linewidth}\centering
        \centering
        \begin{tabular}{l r r r r r r r}
            \toprule            
            Model & Batch size & Dimension & Epochs & Total shot loss & Total area loss & Total loss & Score \\ \hline
            ShuttleNet (Baseline) & 32 & 32 & 150 & 1.3992 & 2.1557 & 3.5549 & 2.826789 \\
            ShuttleNet & 16 & 32 & 300 & 1.0206 & 2.0062 & 3.0268 & 2.692781 \\
            ShuttleNet & 32 & 16 & 300 & 1.1743 & 2.1675 & 3.3418 & 2.565512 \\
            ShuttleNet & 16 & 16 & 300 & 1.1421 & 2.1469 & 3.2890 & 2.470071 \\
            ShuttleNet & 8 & 16 & 300 & 1.0658 & 2.1257 & 3.1915 & 2.526777\\
            ShuttleNet & 16 & 8 & 300 & 1.6032 & 2.2127 & 3.8159 & 2.663351 \\
            Modified ShuttleNet (Ours) & 16 & 16 & 300 & 1.2153 & 2.1732 & 3.3885 & \bf 2.400486 \\
            \bottomrule
        \end{tabular}
        \centering
        \caption{Quantitative results of original ShuttleNet and our modified version on validation set}
        \label{results}
    \end{minipage}
\end{table*}

\begin{figure}[t]
  \begin{minipage}{\columnwidth}
    \includegraphics[width=\textwidth]{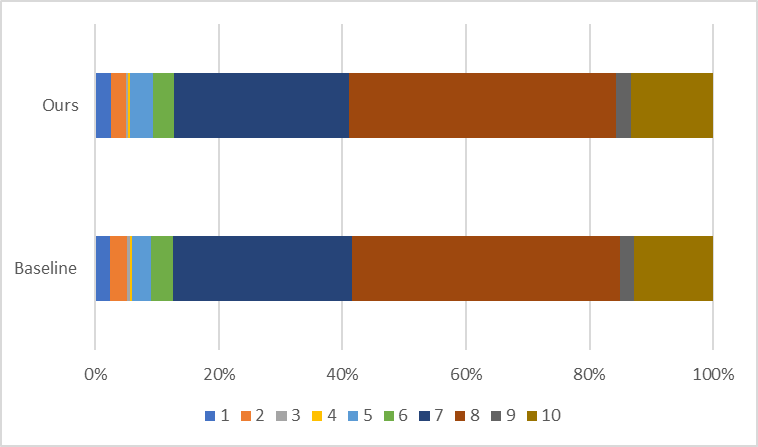}
    \caption*{Figure 4: Predicted landing area zone distributions}
    \label{fig:image5}
  \end{minipage}
  \hfill
  \begin{minipage}{\columnwidth}
  \end{minipage}
\end{figure}

\subsection{Dataset} We utilized the dataset provided on Codalab\footnote{\url{https://codalab.lisn.upsaclay.fr/competitions/12017?secret_key=ba1acc46-1279-4781-94f4-0510fdb017ca}}, which consists of 58 different matches played by 35 players. This dataset includes information such as the order of strokes in a rally, the participating players, and their shot type and area coordinates, which are relevant for our stroke forecasting task.~\cite{ShuttleSet22} 
\nocite{ShuttleSet22}

Figure 1 illustrates that the shot types ``Long Service'' and ``Short Service'' only occur in the first ball round, while ``Defensive Shot'' and ``Drive'' are unlikely to appear in the second ball round. However, the distribution becomes more similar after the third ball round. Conversely, the distribution of shot types based on individual player factors differs significantly, suggesting that each player develops their distinct patterns and tendencies when selecting shots during a match.

Considering the diminishing influence of the first four strokes on subsequent predictions as the match progresses, we removed matches with an excessively high total number of rounds from the training set to prevent our model from overfitting.

\subsection{Improvements based on ShuttleNet} In the embedded layer, we adjust the output for the $i$-th stroke as follows:
\begin{equation}
    e_{i} = \langle e_{i}^s, e_{i}^a \rangle = \langle s'_{i} + p'_{1i}, a'_{i} + p'_{2i} \rangle, 
\end{equation} 
where $s'_{i}, a'_{i}, p'_{1i}, p'_{2i}$ represent the embeddings of shot type, landing area, player number and player location, respectively. 

Firstly, we incorporate the player location embeddings into the landing area embeddings. To make it challenging for their opponent to return the shot, players should choose positions that are farther away from the opponent when executing their attacks. Simultaneously, they need to consider the distance from their current location. These piece of information cannot be determined solely based on the landing coordinates of the shuttlecock.

Secondly, we eliminate the duplicate player embeddings. We believe that in high-intensity matches, players have more control over the shot type they choose, rather than the precise landing spot of the shuttlecock. By removing duplicate player embeddings, we aim to simplify the model and avoid situations where the model predicts a high probability of a drop shot but predicts a far landing spot for the shuttlecock.
As a compromise solution, the personal preference information of the player can still be taken into account during the fusion process. Finally, we removed the ReLU activation function from the area embedding to preserve global information.

\section{Results and Analysis}
\subsection{Experimental Setup}
\paragraph{Implementation details.} We set the dimension of embeddings and contexts to 16. The number of training epochs was 300 and the batch size was 16. The dropout rate in each layer is 0.2. For evaluation in training process, we took 100 samples and choose the closest one to ground truth.

\paragraph{Evaluation Metric.} Cross-entropy (CE) was used to evaluate the results of shot type prediction and mean absolute 
error (MAE) was used to evaluating the results of area coordinate predictions. To reduce the influence of stochasticity, the final score is obtained by selecting the result closest to the ground truth out of the six generated predictive results. The precise formula is described as:

\begin{equation}
    \resizebox{.91\linewidth}{!}{$
        \displaystyle
        l_{i} = \frac{\sum_{r=1}^{\lvert R \rvert} \sum_{n=\tau+1}^{\lvert r \rvert} [S_{n}log\hat{S}_{n}+(\lvert x_{n} - \hat{x}_{n} \rvert + \lvert y_{n} - \hat{y}_{n} \rvert)]}{\lvert R \rvert \cdot (\lvert r \rvert - \tau)}
    $}
\end{equation}

\begin{equation}
    Score = min(l_{1}, l_{2}, l_{3}, l_{4}, l_{5}, l_{6}).
\end{equation}

\paragraph{Baseline.} In this competition, the ShuttleNet  without any architectural or hyperparameter adjustments was chosen as the baseline. ~\cite{ShuttleNet_AAAI_2022}

\subsection{Results and Analysis} Since the competition organizers did not disclose the ground truth of the validation set and the data distribution of the testing set, we could only evaluate our model based on its performance on the training set and the validation set. The results are presented in Table~\ref{results}. Initially, we conducted experiments with hyperparameter adjustments on the original ShuttleNet model to assess its performance. We determined that the best performance was achieved with a batch size of 16, embedding dimension of 16, and training the model for 300 epochs. We also validated that these set of hyperparameters achieved the best performance in our modified model. Increasing the number of epochs beyond 300 may result in a slight decline in performance on the validation set, indicating potential overfitting of the model. Similarly, further reducing the batch size or embedding dimension would cause the model's predictions on the training set to deteriorate rapidly.

\paragraph{Shot type analysis.} In the prediction data, each ball round generates six different predicted outcomes. Figure 2(a) displays the probability distributions of the baseline ShuttleNet and our modified version for each shot type, revealing slight differences between the two models. Additionally, we have plotted the trend of prediction probabilities for both models across different ball rounds in Figure 3. It can be observed that the prediction curve of the baseline ShuttleNet remains relatively stable, indicating consistent predictions. On the other hand, our modified model has learned specific patterns resulting in distinct outcomes. Subsequently, we select the prediction with the highest probability from the same ball round and conduct a voting process to determine the final predicted shot type, as depicted in Figure 2(b). Notably, our model demonstrates a decreased preference for the ``net shot'' and an increased preference for the ``smash'' shot.

\paragraph{Landing area analysis.} Similarly, we would like to compare the prediction of shuttlecock landing spots between the two models. Taking into account the stochastic nature of the predictions and for the sake of convenient comparison, we have converted the predicted landing coordinates into corresponding landing area zones, numbered from 1 to 10, as depicted in Figure 4. It is evident that both models demonstrate similar distributions, suggesting comparable accuracy in predicting the landing areas.

\section{Conclusion} In this paper, we analyzed the training set of the CoachAI Badminton Challenge and proposed a modified approach to adapt the ShuttleNet model for this task. We found that improving the prediction of shuttlecock landing spots based on ShuttleNet was challenging, while there was significant room for improvement in predicting the shot types.

\bibliographystyle{named}
% \bibliography{ijcai23}

\end{document}